\pdfoutput=1
\documentclass[11pt]{article}
\usepackage{acl}
\usepackage{times}
\usepackage{latexsym}
\usepackage{graphicx}
\usepackage{caption}
\usepackage{subcaption}
\usepackage[T1]{fontenc}
\usepackage[utf8]{inputenc}
\usepackage{microtype}
\usepackage{multirow}
\usepackage{amsmath}
\usepackage{amssymb}
\usepackage{amsthm}
\usepackage{adjustbox}
\usepackage{makecell}
\usepackage{arydshln}
\usepackage{verbatim}
\usepackage[colorinlistoftodos]{todonotes}
\usepackage[normalem]{ulem}
\usepackage{enumitem}

\DeclareMathOperator{\KL}{KL}

\DeclareMathOperator{\Softmax}{\sigma}

\DeclareMathOperator{\RL}{RL}
\DeclareMathOperator{\TL}{TL}
\DeclareMathOperator{\BoW}{BoW}
\DeclareMathOperator{\negt}{neg}

\DeclareMathOperator{\RBO}{RBO}
\DeclareMathOperator{\IRBO}{IRBO}

\renewcommand{\vec}[1]{\boldsymbol{\mathbf{#1}}}

\title{Improving Contextualized Topic Models with Negative Sampling}

\author{Suman Adhya,  Avishek Lahiri, Debarshi Kumar Sanyal \\
         Indian Association for the Cultivation of Science, Jadavpur, Kolkata-700032, India \\
         \{ \texttt{adhyasuman30}, \texttt{avisheklahiri2014}, \texttt{debarshisanyal} \}\texttt{@gmail.com} 
 \AND        
    Partha Pratim Das \\
  Ashoka University, Sonipat, Haryana-131029, India \\
  Indian Institute of Technology Kharagpur, West Bengal-721302, India \\
  \texttt{ppd@cse.iitkgp.ac.in}
}

\begin{document}
\maketitle
\begin{abstract}
Topic modeling has emerged as a dominant method for exploring   large document collections. Recent approaches to topic modeling use large contextualized language models and variational autoencoders. In this paper, we propose a negative sampling mechanism for a contextualized topic model to improve the quality of the generated topics. In particular, during model training, we perturb the generated document-topic vector and use a triplet loss to encourage the document reconstructed from the correct document-topic vector to be similar to the input document and dissimilar to the document reconstructed from the perturbed vector. Experiments for different topic counts on three publicly available benchmark datasets show that in most cases, our approach leads to an increase in topic coherence over  that of the baselines. Our model also achieves very high topic diversity. 
\end{abstract}

\section{Introduction}
The modern world is witnessing tremendous growth in digital documents. It is often necessary to organize them into semantic categories to make the content more easily accessible to users.  The assignment of domain tags through manual intervention can be quite cumbersome and very expensive to maintain, mainly due to the enormity and diversity of the available data. The use of topic modelling techniques can be of huge significance in this area because of their ability to automatically learn the overarching themes or topics from a collection of documents in an unsupervised way and tag the documents with their dominant topics \cite{newman_jcdl, boyd2017applications, adhya-sanyal-2022-indian}. Informally, a topic is a group of extremely related words. 
While latent Dirichlet allocation (LDA) \cite{blei2003latent} is the classical topic modeling approach, recently neural topic models have become popular as they decouple the inference mechanism from the underlying modeling assumptions (e.g., the topic prior), thereby simplifying the design of new topic models. Neural topic models are based on variational autoencoders (VAEs) \cite{kingma2014auto} and allow us to leverage the progress in deep learning in modeling text \cite{zhaotopic}. The recently proposed contextualized topic model (CTM)  \cite{bianchi-etal-2021-pre}, which is a neural topic model,  represents each document in the collection both as a bag-of-words (BoW) vector as well as a dense  vector produced by a pre-trained transformer like sentence-BERT (SBERT) \cite{reimers-gurevych-2019-sentence}, thus combining a classical representation with a contextualized representation that captures the semantics of the text better. CTM produces state-of-the-art performance on many benchmark datasets  \cite{bianchi-etal-2021-pre}. 

A neural topic model is trained to maximize the log-likelihood of the reconstruction of the input document  and minimize the KL-divergence of the learned distribution of the latent (topic) space from a known prior distribution  of the latent space. 
If the topics in a document are perturbed, that is, say, the top topic in a document is deleted, the document should display a marked change in its word distribution. Such an objective is not explicitly modeled above. In this paper, we train CTM to infer topics from a document in such a way that while the inferred topics should aid in reconstructing the document (as in any topic modeling algorithm), when the top topics are perturbed it should fail to reconstruct the original document. This is done by treating the document reconstructed from the correct topic vector as an anchor that is encouraged to be similar to the original input document but dissimilar to the document reconstructed from the perturbed topics. Our proposed model, \textbf{CTM-Neg},  achieves higher average topic coherence, measured by NPMI score, than that of other competing topic models, and very high topic diversity on three datasets. We have made our code publicly available\footnote{\url{https://github.com/AdhyaSuman/CTMNeg}}.

Thus, our primary contributions are: 
\begin{enumerate}
    \item We propose a \textit{simple but effective negative sampling technique for  neural topic models}. Negative samples are produced automatically in an unsupervised way.
    \item We perform extensive experiments on three publicly available datasets. In particular, we compare the proposed model with four other topic models for eight different topic counts on each dataset. We observe that the proposed strategy \textit{leads to an increase in topic coherence} over the baselines in most of the cases.
    Averaged over different topic counts, CTM-Neg achieves the highest mean NPMI score on all three datasets, and highest mean CV on two datasets, and the second-highest mean CV on the third. CTM-Neg also attains the best or the second best mean topic diversity scores on the three datasets though all the topic models except one (which underperforms) produce similar high topic diversity.
\end{enumerate} 

\section{Related Work}
\label{sec:relatedWork}
Latent Dirichlet allocation (LDA) \cite{blei2003latent} models every document in a given corpus as a mixture of topics, where each topic is a probability distribution over the vocabulary. Among the modern neural alternatives to LDA, a pioneering approach is the ProdLDA model \cite{srivastava2017autoencoding}. It is a VAE-based topic model that uses an approximate Dirichlet prior (more precisely, the Laplace approximation to the Dirichlet prior in the softmax basis), instead of a standard Gaussian prior \cite{miao2016neural}. The VAE takes a bag-of-words (BoW) representation of a document, maps it to a latent vector using an encoder or inference network, and then maps the vector back to a discrete distribution over words using a decoder or generator network.  CTM \cite{bianchi-etal-2021-pre} augments ProdLDA by allowing in its input a contextualized representation (SBERT) of the  documents. Embedded topic model (ETM) \cite{dieng-etal-2020-topic} is a VAE-based topic model that uses distributed representations of both words and topics.

Negative sampling in NLP-based tasks was popularized after its use in the  word embedding model, word2vec \cite{Mikolov}. The idea of negative sampling is to `sample' examples from a noise distribution and ensure that the model being trained can distinguish between the positive and negative examples. It can be used to reduce the computational cost of training, help identify out-of-distribution examples, or to make the model more robust to adversarial attacks \cite{xu2022negative}.  
A few works have recently applied it to topic modeling.  For example, \cite{wu2020short} proposed a negative sampling and quantization model (NQTM) with a modified cross-entropy loss to generate sharper topic distributions from short texts. Some researchers have applied generative adversarial networks to design topic models \cite{wang2019atm, hu2020neural, BAT}, but since the negative examples are generated from an assumed fake distribution, they bear little similarity to real documents. 
In \cite{nguyen2021contrastive}, a negative document sample is created by replacing the weights of the words having the highest tf-idf scores in the input document with the weights of the same words in the reconstructed document. Our method follows a different strategy: it  generates a perturbed document-topic vector (instead of an explicit negative document) and uses triplet loss to push the BoW vector reconstructed from the correct topic vector closer to the input BoW vector and farther from the BoW vector generated from the perturbed topics. Unlike the present work, none of the other adversarial topic models use contextual embeddings as input.

\begin{figure*}[!htbp]
   \centering
   \includegraphics[width=\linewidth]{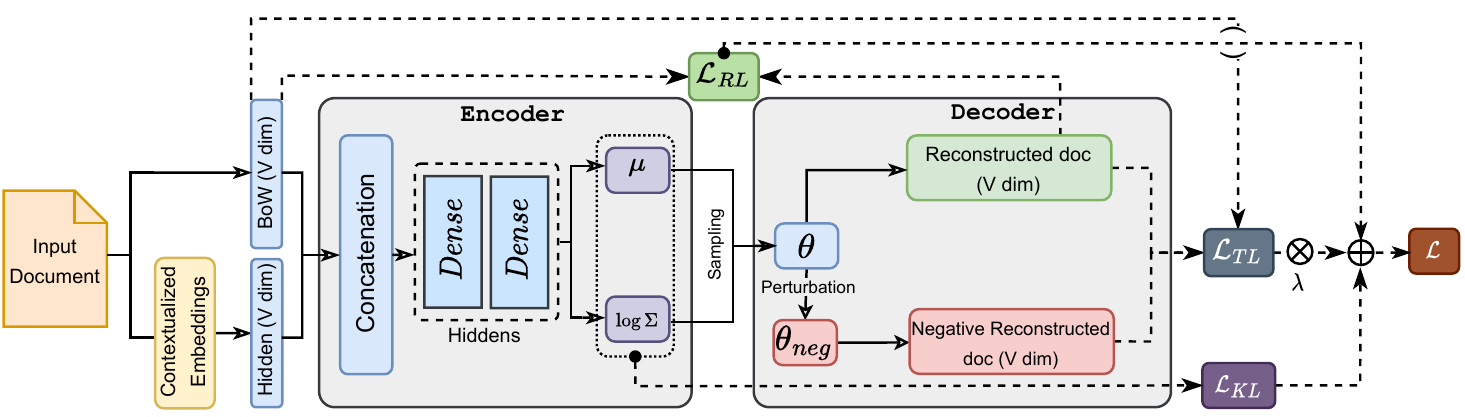}
   \caption{Framework for the contextualized topic model with negative sampling (CTM-Neg).}
   \label{fig:model_arch}
\end{figure*}

\section{Proposed Method}
\label{sec:proposedMethod}
\subsection{Baseline Architecture}
\label{sec:genArch}
Our proposed model is based on a VAE architecture. In particular, we build upon CTM  \cite{bianchi-etal-2021-pre}. We assume that the vocabulary size is $V$ and a document is represented as a normalized bag-of-words vector $\vec{x}_{\BoW}$ as well as a contextualized embedding vector $\vec{x}_c$.  
A linear layer converts $\vec{x}_c$ to a $V$-dimensional vector. The encoder of the VAE concatenates these two vectors into a single $2V$-dimensional vector $\vec{x}$  and outputs the parameters of the posterior $\left( \vec{\mu}_{T \times 1},  \vec{\Sigma}_{T \times 1} \right)$ where $T$ is the number of topics, $\vec{\mu}_{T \times 1}$ denotes the mean, and  $\vec{\Sigma}_{T \times 1}$ represents the diagonal covariance matrix. Note that it is standard in the VAE literature to assume a diagonal covariance matrix instead of a full covariance matrix \cite{srivastava2017autoencoding}.
In the decoder, using the reparameterization trick the latent representation $(\vec{z}_{T \times 1})$ is generated:
\begin{align}
    \vec{z}_{T \times 1} &= \vec{\mu}_{T \times 1} + \vec{\Sigma}_{T \times 1}^{1/2} \odot \vec{\epsilon}_{T \times 1}
     \label{eq:repar}
\end{align}
where $\vec{\epsilon}_{T \times 1} \sim \mathcal{N}(\vec{0}, \vec{I})$ and $\odot$ denotes Hadamard product. 
This hidden representation $\vec{z}$ is then used as a logit of a softmax function (${\sigma(\cdot)}$) to generate the document-topic distribution $\vec{\theta}_{T \times 1}$ ($= \sigma(\vec{z}_{T \times 1})$).
The decoder has an unnormalized topic-word matrix $\vec{\beta}_{T \times V}$, which is used to reconstruct the word distribution in the following manner:
\begin{align}
    \hat{\vec{x}}_{V \times 1} = \Softmax ( \vec{\beta}_{T \times V}^{\top} {\vec{\theta}}_{T \times 1}   )
     \label{eq:anchor}
\end{align}

To formulate the loss function, note that the encoder learns the posterior distribution $q_{\vec{\phi}}(\vec{z}|\vec{x})$.
We assume that the prior is $p(\vec{z})$. The decoder is the generative model $p_{\vec{\theta}}(\vec{x}_{\BoW}|\vec{z})$. 
The loss function to be minimized is given by
\begin{align}
   \label{eq:loss1}
    \mathcal{L}_{\operatorname{CTM}} &= \mathcal{L}_{\RL} + \mathcal{L}_{\KL} \nonumber \\
    &\equiv - \mathbb{E}_{\vec{z}
    \sim q_{\vec{\phi}}(\vec{z}|\vec{x})} p_{\vec{\theta}}(\vec{x}_{\BoW}|\vec{z}) \nonumber \\
    & \qquad \qquad + \operatorname{D}_{\KL}\left(q_{\vec{\phi}}(\vec{z}|\vec{x})||p(\vec{z})\right)
\end{align}
here, the first term ($\mathcal{L}_{\RL}$) is the reconstruction loss (measured by the cross-entropy between the predicted output distribution $\hat{\vec{x}}$ and the input vector $\vec{x}_{\BoW}$) while the second term $\mathcal{L}_{\KL}$ is the KL-divergence of the learned latent space distribution $q_{\vec{\phi}}(\vec{z}|\vec{x})$ from the prior $p(\vec{z})$ of the latent space. 

\subsection{Proposed Negative Sampling Mechanism}
To improve the topic quality, we train the above model with negative samples as follows. For every input document, after a topic  vector $\vec{\theta}$ is sampled, a perturbed vector $\tilde{\vec{\theta}}_{\negt}$ is generated from it by setting the entries for the top $S$ topics (i.e., the $S$ positions in $\vec{\theta}$ corresponding to the $S$ largest values in $\vec{\theta}$) to zero. $\tilde{\vec{\theta}}_{\negt}$ is then normalized so that the resulting  vector $\vec{\theta}_{\negt}$ is a probability vector. The normalization is done simply by  dividing the values in $\tilde{\vec{\theta}}_{\negt}$ by their sum, as all values, in $\tilde{\vec{\theta}}_{\negt}$ are already non-negative (since $\vec{\theta}$ is obtained by applying \texttt{softmax}).
Mathematically,
\begin{align}
    \vec{\theta}_{\negt} &= \frac{\tilde{\vec{\theta}}_{\negt}}{\sum_{i=1}^T \tilde{{\theta}}_{\negt}[i]}  \\
    \text{ where, } \tilde{{\theta}}_{\negt}[i]&= 
    \begin{cases}
      0 & \text{if } i \in \operatorname{argmax}(\vec{\theta},S) \\
      {\theta}[i] & \text{otherwise}
    \end{cases} \nonumber 
\end{align}
The function $\texttt{argmax}(\vec{\theta},S)$ returns the indices of the $S$ largest values in  $\vec{\theta}$. We treat $S$ as a hyperparameter. 
Like $\vec{\theta}$, the perturbed topic vector $\vec{\theta}_{\negt}$ is passed through the decoder network. The latter generates $\hat{\vec{x}}_{\negt} = \Softmax(\vec{\beta}^\top \vec{\theta}_{\negt})$. 
We introduce a new term, triplet loss $\mathcal{L}_{\TL}$, in Eq. \eqref{eq:loss1} assuming the anchor is $\hat{\vec{x}}$, the positive sample is $\vec{x}_{\BoW}$ (the original input document), and the negative sample is $\hat{\vec{x}}_{\negt}$:
\begin{equation}
    \mathcal{L}_{\TL} = \max(||\hat{\vec{x}} - \vec{x}_{\BoW}||_2 - ||\hat{\vec{x}} - \hat{\vec{x}}_{\negt}||_2 + m, 0 ) 
    \label{eq:tloss}
\end{equation}
where $m$ is the margin. 
Therefore, the modified loss function to be minimized is given by:
\begin{align} \label{eq:loss2}
    \mathcal{L} &= \left(\mathcal{L}_{\RL} + \mathcal{L}_{\KL}\right) + \lambda \mathcal{L}_{\TL}
\end{align}
where $\lambda$ is a hyperparameter. Fig. \ref{fig:model_arch} depicts the proposed model. The model is trained in an end-to-end manner using Adam optimizer and backpropagation.

\begin{table*}[!htbp]
    \centering
    \begin{adjustbox}{width=0.98\linewidth}
    \begin{tabular}{|c|c|c|c|c|c|c|c|c|}
        \hline
         \multirow{ 2}{*}{\textbf{Dataset}} &  \multicolumn{8}{c|}{\textbf{\#Topics}} \\ \cline{2-9}
          & \textbf{10} & \textbf{20} & \textbf{30} & \textbf{40} & \textbf{50} & \textbf{60} & \textbf{90} & \textbf{120} \\ \hline
         GN & (2, 0.7) & (2, 0.58) & (2, 0.59) & (2, 0.59) & (3, 0.82) & (3, 0.94) & (1, 0.68) & (3, 0.82) \\
         20NG & (3, 0.78) & (3, 0.83) & (3, 0.86) & (1, 0.74) & (1, 0.12) & (3, 0.27)  & (1, 0.84) & (1, 0.90)\\
         M10 & (3, 0.9) & (3, 0.49) & (1, 0.82) & (1, 0.59) & (3, 0.82)   & (3, 0.58)  & (3, 0.93) & (3, 0.27) \\ \hline
    \end{tabular}
    \end{adjustbox}
    \caption{Each paired entry shows the best hyperparameters $(S, \lambda)$ in CTM-Neg as discovered by OCTIS for a given (Dataset, \#Topics) combination.}
    \label{tab:hparam}
\end{table*}

\section{Experimental Setup}
\label{sec:experimentalSetup}
We perform all experiments in \href{https://github.com/MIND-Lab/OCTIS}{OCTIS}  \cite{terragni2020octis}, which is an integrated framework for topic modeling.
\subsection{Datasets}
We use the following three datasets: 
\begin{enumerate}
    \item \textbf{GoogleNews} (\textbf{GN}): It consists of $11,109$ news articles, titles, and snippets collected from the Google News website in November 2013 \cite{GNdataset}.
    \item \textbf{20NewsGroups} (\textbf{20NG}): It comprises $16,309$ newsgroup documents partitioned (nearly) evenly across 20 different newsgroups \cite{terragni2020octis}.
    \item \textbf{M10}: It is a subset of CiteSeer$^\text{X}$ data comprising 8355 scientific  publications from 10 distinct research areas \cite{10.5555/3060832.3060886}.
\end{enumerate}
The last two datasets are available in OCTIS while we added the first one.
\subsection{Evaluation Metrics}
Coherence measures help to assess the relatedness between the top words of a topic. Informally, a topic is said to be coherent if it contains words that, when viewed together, help humans to recognize it as a distinct category \cite{hoyle2021automated}.  
We use \textbf{Normalized Pointwise Mutual Information} (\textbf{NPMI}) and 
\cite{lau-etal-2014-machine} and \textbf{Coherence Value} (\textbf{CV}) \cite{roder2015exploring} to measure topic coherence. 
NPMI is widely adopted as a proxy for human judgement of topic coherence though some researchers also use CV (but CV has some known \href{https://palmetto.demos.dice-research.org/}{issues}). 
NPMI calculates topic coherence by measuring how likely the topic words are to co-occur. 
If $p(w_i, w_j)$ represents the probability of two words $w_i$ and $w_j$ co-occurring in a boolean sliding context window, and $p(w_i)$ is the marginal probability of word $w_i$, then the NPMI score is given by \cite{lau-etal-2014-machine},
    \begin{equation}
	\operatorname{NPMI}(w_i, w_j) = { \left( \frac{\log \frac{p(w_i, w_j) + \epsilon}{p(w_i).p(w_j)}}{- \log (p(w_i, w_j) + \epsilon)} \right) }
    \end{equation}
where $\epsilon$ is a small positive constant used to avoid zero.
$\operatorname{NPMI}(w_i, w_j)$ lies in $[-1, \: +1]$ where $-1$ indicates the words never co-occur and $+1$ indicates they always co-occur. 
CV is calculated using an indirect cosine measure along with the NPMI score over a boolean sliding window 
\cite{roder2015exploring, krasnashchok2018improving}. OCTIS uses the \texttt{CoherenceModel} of \href{https://radimrehurek.com/gensim/models/coherencemodel.html}{gensim} where NPMI is referred to as \texttt{c\_npmi} and CV as \texttt{c\_v}.  

We measure the diversity of topics using \textbf{Inversed Rank-Biased Overlap} (\textbf{IRBO}) \cite{bianchi-etal-2021-pre}. It gives $0$ for identical topics and $1$ for completely dissimilar topics. Suppose we are given a collection $\aleph$ of $T$ topics where each topic is a list of words such that the words at the beginning of the list have a higher probability of occurrence (i.e., are more important or more highly ranked) in the topic. Then, the IRBO score of the topics is defined as
\begin{align}
    \IRBO(\aleph) = 1 - \frac{\sum_{i=2}^T \sum_{j=1}^{i-1} \RBO(l_i, l_j)}{n} 
\end{align}
where $n = {\binom{T}{2}}$ is the number of pairs of lists, and $\RBO(l_i, l_j)$ denotes the standard Rank-Biased Overlap between two ranked lists $l_i$ and $l_j$ \cite{webber2010similarity}. IRBO allows the comparison of lists that may not contain the same items, and in particular, may not cover all items in the domain. Two lists (topics) with overlapping words receive a smaller IRBO score when the overlap occurs at the highest ranks of the lists than when they occur at lower ranks. IRBO is implemented in OCTIS. Higher values of NPMI, CV, and IRBO are better than lower values.

In our experiments, for evaluation using the above metrics in OCTIS, we use the top-10 words from every topic and the default values for all the other parameters.

\subsection{Baselines and Configuration}
 We denote our proposed topic model by \textbf{CTM-Neg}. As baselines we use the following topic models, which are already implemented in OCTIS:
 \begin{enumerate}
     \item \textbf{CTM} \cite{bianchi-etal-2021-pre}.
     \item \textbf{ProdLDA} \cite{srivastava2017autoencoding}.
     \item \textbf{ETM} \cite{dieng-etal-2020-topic}.
     \item \textbf{LDA} \cite{blei2003latent}.
 \end{enumerate}

In CTM-Neg, CTM, and Prod-LDA, the encoder is a fully-connected feedforward neural network (FFNN) with  two hidden layers with 100 neurons each, and the decoder is a single-layer FFNN.
We use {\href{https://huggingface.co/sentence-transformers/paraphrase-distilroberta-base-v2}{\texttt{paraphrase-distilroberta-base-v2}}} (which is an SBERT  model) to obtain the contextualized representations of the input documents in CTM and CTM-Neg.

In CTM-Neg, we set $m=1$ in Eq. \eqref{eq:tloss} as is the \href{https://pytorch.org/docs/stable/generated/torch.nn.TripletMarginLoss.html}{default in PyTorch}. We have optimized the hyperparameters $S$ and $\lambda$ using the Bayesian optimization framework of OCTIS to maximize NPMI. Table \ref{tab:hparam} shows the optimal  values discovered when $S \in \{1,2,3\}$ and $\lambda \in [0,1]$. In LDA, we use 5 passes over the input corpus as the default single pass produces too poor topics. 
Other hyperparameters are set to their default values in OCTIS.
For all datasets, the vocabulary is set to the most common 2K words in the corpus.
Experiments for each topic model are done for all topic counts in the set $\{10,20,30,40,50,60,90,120\}$.
We have trained all models for 50  epochs. 

\begin{table*}[!htbp] 
    \centering
    \begin{adjustbox}{width=.9\linewidth}
    \begin{tabular}{|c|c|cc | cc|cc|}
    \hline
         \multirow{3}{*}{\textbf{Dataset}} & \multirow{3}{*}{\textbf{Model}} & \multicolumn{4}{c|}{\textbf{Coherence}} & \multicolumn{2}{c|}{\textbf{Diversity}} \\
         & & \multicolumn{2}{c}{\textbf{NPMI}}  & \multicolumn{2}{c|}{\textbf{CV}} & \multicolumn{2}{c|}{\textbf{IRBO}} \\ \cline{3-8}
         & & Mean & Median & Mean & Median & Mean & Median \\ \hline
    
        \multirow{5}{*}{\textbf{GN}} & CTM-Neg & \makecell{\textbf{0.142}} & \makecell{\textbf{0.188}} & \makecell{\textbf{0.530}} & \makecell{\textbf{0.552}} & \makecell{\textbf{0.998}} & \makecell{\textbf{0.998}} \\ 
        
        & CTM & \makecell{0.081} & \makecell{0.128} & \makecell{0.485} & \makecell{0.513} & \makecell{0.995} & \makecell{0.995} \\ 
        
        & ProdLDA & \makecell{0.056} & \makecell{0.076} & \makecell{0.471} & \makecell{0.476} & \makecell{0.996} & \makecell{0.996} \\ 
                
        & ETM & \makecell{-0.263} & \makecell{-0.271} & \makecell{0.414} & \makecell{0.416} & \makecell{0.627} & \makecell{0.660} \\
        
        & LDA & \makecell{-0.164} & \makecell{-0.176} & \makecell{0.403} & \makecell{0.405} & \makecell{0.997} & \makecell{\textbf{0.998}} \\ \hline

        \multirow{5}{*}{\textbf{20NG}} & CTM-Neg  & \makecell{\textbf{0.121}} & \makecell{\textbf{0.127}} & \makecell{\textbf{0.648}} & \makecell{\textbf{0.653}} & \makecell{\textbf{0.991}} & \makecell{\textbf{0.991}} \\ 
        
        & CTM & \makecell{0.093} & \makecell{0.098} & \makecell{0.627} & \makecell{0.632} & \makecell{0.990} & \makecell{0.990} \\ 
        
        & ProdLDA & \makecell{0.080} & \makecell{0.084} & \makecell{0.609} & \makecell{0.607} & \makecell{0.990} & \makecell{\textbf{0.991}} \\ 
                
        & ETM & \makecell{0.049} & \makecell{0.048} & \makecell{0.528} & \makecell{0.527} & \makecell{0.819} & \makecell{0.808} \\
        
        & LDA & \makecell{0.075} & \makecell{0.080} & \makecell{0.571} & \makecell{0.577} & \makecell{0.983} & \makecell{0.990} \\ \hline

      \multirow{5}{*}{\textbf{M10}} & CTM-Neg & \makecell{\textbf{0.052}} & \makecell{\textbf{0.056}} & \makecell{0.462} & \makecell{\textbf{0.461}} & \makecell{0.986} & \makecell{0.985} \\ 
        
        & CTM & \makecell{0.048} & \makecell{0.047} & \makecell{\textbf{0.466}} & \makecell{\textbf{0.461}} & \makecell{0.980} & \makecell{0.979} \\ 
        
        & ProdLDA & \makecell{0.025} & \makecell{0.023} & \makecell{0.448} & \makecell{0.449} & \makecell{0.983} & \makecell{0.981} \\ 
                
        & ETM & \makecell{-0.056} & \makecell{-0.062} & \makecell{0.345} & \makecell{0.350} & \makecell{0.502} & \makecell{0.484} \\
        
        & LDA & \makecell{-0.192} & \makecell{-0.201} & \makecell{0.386} & \makecell{0.389} & \makecell{\textbf{0.989}} & \makecell{\textbf{0.992}} \\ \hline

    \end{tabular}
    \end{adjustbox}
    \caption{Comparison of topic models on three datasets. For each metric and each topic model, we mention the mean and the median of the scores for topic counts \{10, 20, 30, 40, 50, 60, 90, 120\}.
    \label{tab:compareTM2}
    }
\end{table*}

\begin{figure*}[!htbp]
\centering
\begin{subfigure}[b]{\textwidth}
   \includegraphics[width=1\linewidth]{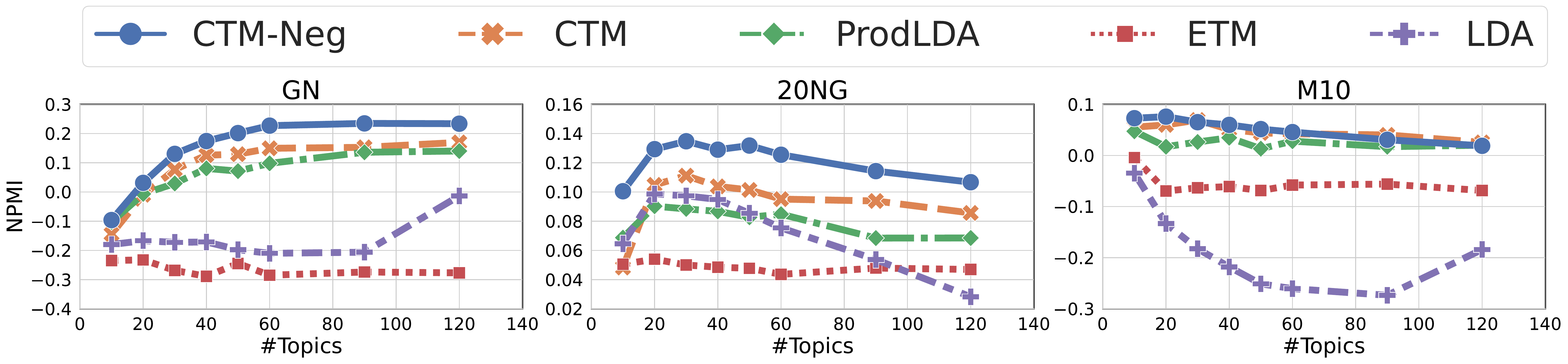}
   \label{fig:NPMI} 
\end{subfigure}

\begin{subfigure}[b]{\textwidth}
   \includegraphics[width=1\linewidth]{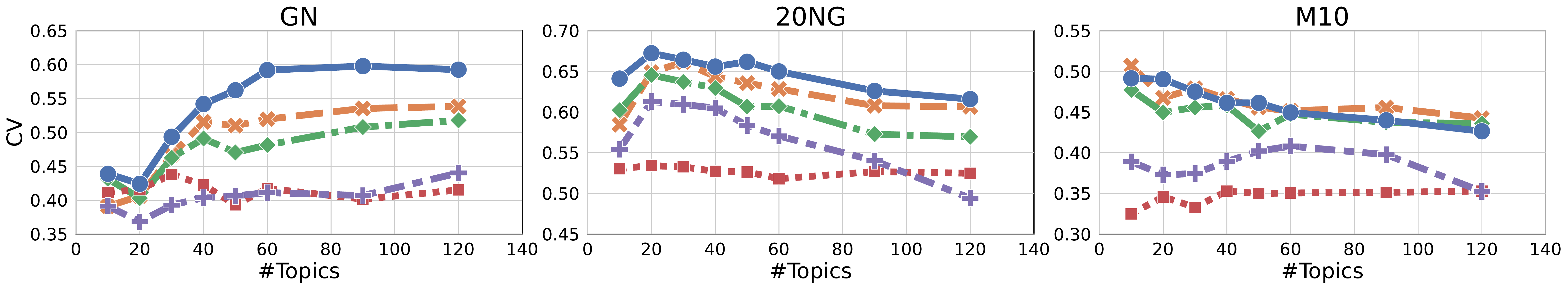}
   \label{fig:CV}
\end{subfigure}

\begin{subfigure}[b]{\textwidth}
   \includegraphics[width=1\linewidth]{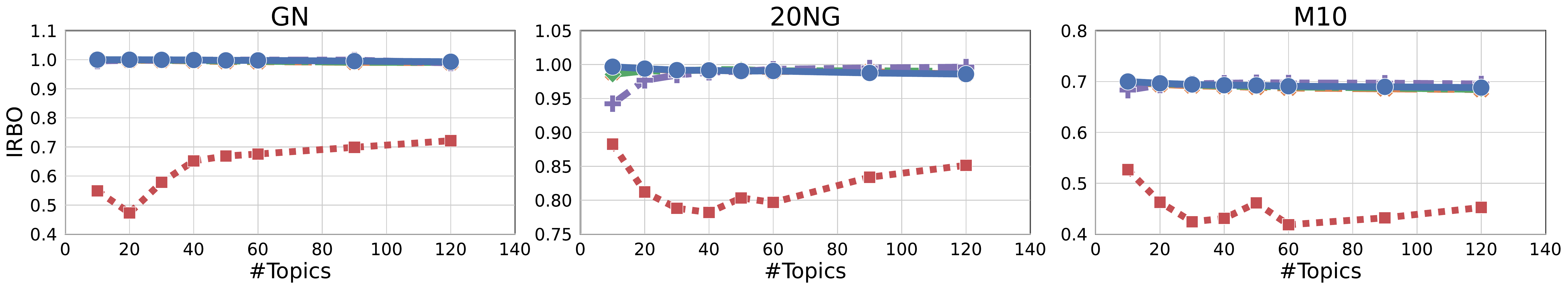}
   \label{fig:IRBO}
\end{subfigure}
\caption{Variation of topic coherence (NPMI and CV) and topic diversity (IRBO)  with topic count for different topic models on three  datasets. The ordinate value of each data point reports the median over five independent runs.
}
\label{fig:compareTM}
\end{figure*}

\section{Results}
\label{sec:results}
\subsection{Quantitative Evaluation}
Given a dataset and a topic model, we recorded the median values of NPMI, CV, and IRBO over 5 independent runs for each topic count. We choose median instead of mean as the former is more robust to outliers. Then for the same dataset and topic model, we compute the average of these values so that we can get an idea of the performance of the topic model without coupling it to a specific topic count. Table \ref{tab:compareTM2} shows the corresponding values where we mention the median along with the mean.

We observe that CTM-Neg achieves the highest average NPMI on all datasets. CTM-Neg also produces the highest average CV on all datasets except M10 where CTM performs slightly better. In the case of IRBO, while CTM-Neg gives the highest scores on GN and 20NG, it ranks as the second best on M10. It is also observed that the IRBO values for all models except ETM are very close to each other. 

In order to afford a more fine-grained view of the performance of the topic models, Fig. \ref{fig:compareTM} depicts how the scores vary with topic count for all topic models and on all datasets. CTM-Neg always achieves the highest NPMI and CV scores on GN and 20NG datasets. On the M10 corpus, CTM scores slightly better than CTM-Neg in NPMI and CV for some topic counts.
The IRBO plots in Fig. \ref{fig:compareTM} show that on a given dataset, all topic models, except ETM, achieve very similar IRBO scores for every topic count. ETM is always found to produce significantly lower IRBO values. CTM-Neg does not always produce the highest IRBO. For example, on the M10 corpus, the IRBO score of CTM-Neg is the highest till $T=20$ after which LDA dominates and CTM-Neg is relegated to the second position. A closer look at Fig. \ref{fig:compareTM} reveals that this gain in topic diversity for LDA comes at the expense of reduced NPMI. 

\begin{table}[!htbp]
\centering
\begin{adjustbox}{width=.9\linewidth}
  \begin{tabular}{ | c | c | } \hline
    \textbf{Label} & \textbf{\#Documents} \\ \hline
    Agriculture & 643 \\
    Archaeology & 131 \\
    Biology & 1059 \\
    Computer Science & 1127 \\
    Financial Economics	& 978 \\
    Industrial Engineering & 944 \\
    Material Science & 873 \\
    Petroleum Chemistry	& 886 \\
    Physics	& 717 \\
    Social Science & 997 \\ \hline
\end{tabular}
\end{adjustbox}
\caption{M10 labels with corresponding document counts.\label{tab:M10_labels}}
\end{table}

\begin{figure}[!htbp]
    \centering
    \includegraphics[width=1\linewidth]{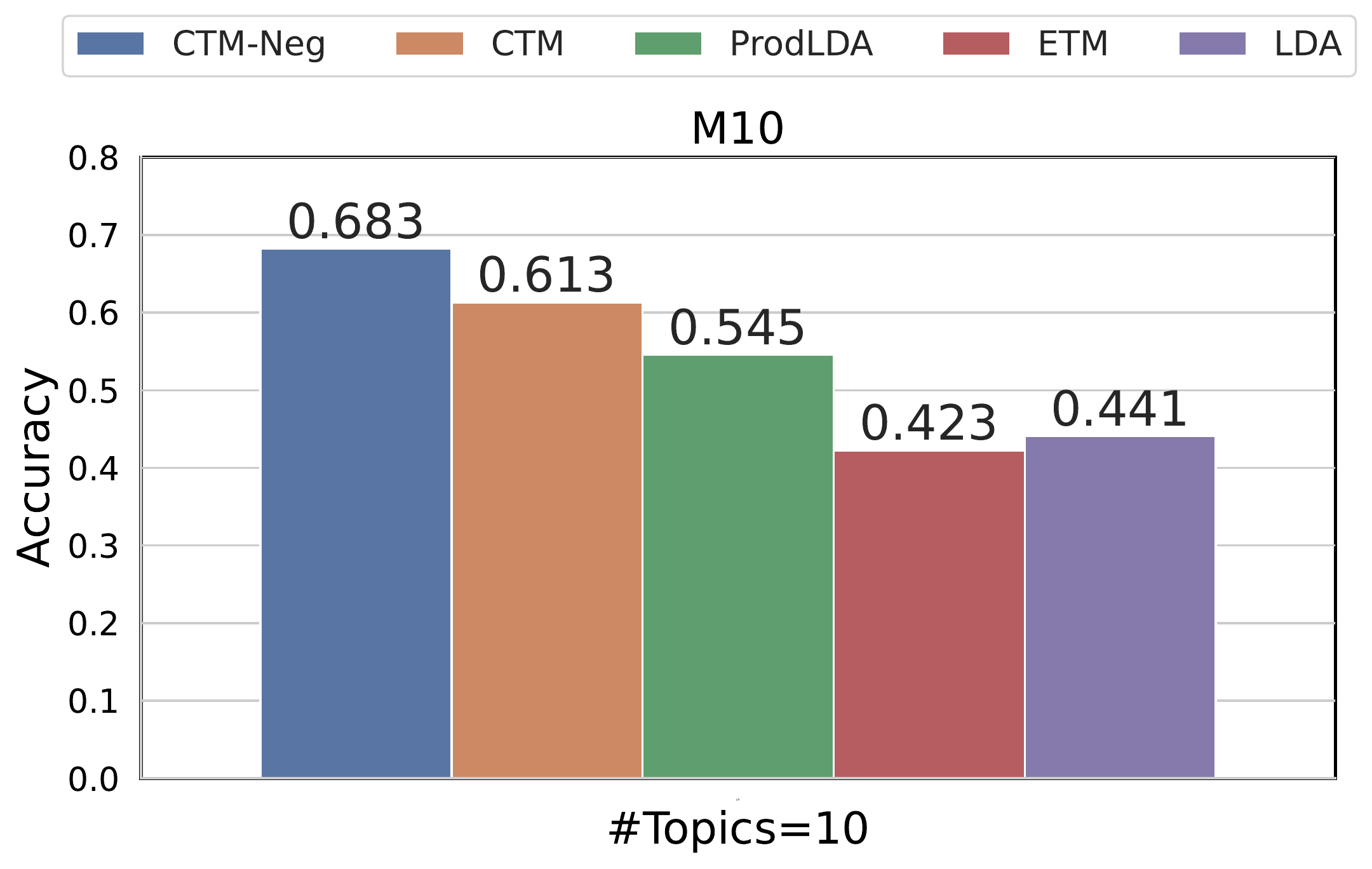}
    \caption{Document classification for M10 corpus with $T=10$ topics.}
    \label{fig:docClass}
\end{figure}

\subsection{Extrinsic Evaluation}
We also use an extrinsic task to evaluate the topic models. We measure the predictive performance of the generated topics on a document classification task. Specifically, we use the M10 dataset from OCTIS where each document is already marked with one of 10 class labels as shown in Table \ref{tab:M10_labels}.
The corpus is divided into train/dev/test subsets in the ratio 70:15:15. Each topic model is trained on the training subset to produce $T=10$ topics and the   $T$-dimensional document-topic latent vector is used as a representation of the document. Next, a linear support vector machine is trained with these representations of the training subset (for each topic model), and the performance on the test subset is recorded. Fig. \ref{fig:docClass} shows that CTM-Neg achieves the highest accuracy.

\begin{table*}[!htbp]
\centering
\begin{adjustbox}{width=1.0\linewidth}
  \begin{tabular}{ | c | c | } \hline
    \textbf{Model} & \textbf{Topics} \\ \hline
     \multirow{4}{*}{CTM-Neg} 
                             & turkish, armenian, jewish, population, muslim, village, israeli, genocide, government, war \\
                             & chip, key, encryption, government, clipper,  phone, security, privacy, escrow, secure \\
                             & video, monitor, vga, port, modem, apple, driver, card, resolution, board  \\ 
                             & score, playoff, period, play, fan, win, hockey, game, baseball, lose \\
                             \hline
     \multirow{4}{*}{CTM} 
        & people, armenian, soldier, village, turkish, massacre, troop, neighbor, town, genocide \\
        &  chip, clipper, encryption, government, encrypt, algorithm, agency, secure, phone, key\\
        & draw, mouse, advance, convert, font,  screen, button, host, code, terminal \\
        & game, win, final, goal, period, cap, score, fan, lead, play \\ \hline

     \multirow{4}{*}{ProdLDA} 
        & genocide, armenian, turkish, greek, muslim, village, population, russian, massacre, jewish \\
        & encryption, secret, secure, chip, privacy, government, key, agency, security, encrypt \\
        & monitor, card, apple, video, sale, price, board, audio, offer, external \\
        & game, team, division, season, hockey, playoff, score, goal, player, wing \\ \hline
     \multirow{4}{*}{ETM} 
        & people, kill, child, gun, armenian, fire, man, time, leave, start \\
        & key, chip, encryption, clipper, bit, government, algorithm, message, law, system \\
        & drive, card, disk, system, bit, run, window, scsi, driver, monitor \\
        & game, play, win, team, player, year, good, score, hit, season \\ \hline
     \multirow{4}{*}{LDA} 
        & people, jewish, armenian, child, man, kill, woman, death, turkish, israeli \\
        & key, chip, encryption, government, security, clipper, bit, public, message, system \\
        & card, work, monitor, system, driver, problem, run, machine, video, memory \\
        & game, team, play, player, win, year, good, season, hit, score \\ \hline

\end{tabular}
\end{adjustbox}
\caption{Some related topics discovered by different topic models in the 20NG corpus when run for $T=20$ topics. \label{tab:topics_20NG}}
\end{table*}

\begin{table*}[!htbp]
\centering
\begin{adjustbox}{width=1.0\linewidth}
  \begin{tabular}{ | c | c | } \hline
    \textbf{Model} & \textbf{Topics} \\ \hline
     \multirow{4}{*}{CTM-Neg} 
         & neural, network, learn, recurrent, learning, artificial, language, evolutionary, genetic, adaptive \\ 
         & expression, gene, datum, sequence, cluster, protein, microarray, dna, analysis, motif \\ 
         & stock, return, market, price, volatility, exchange, rate, interest, option, monetary \\ 
         & decision, make, agent, making, group, multi, uncertainty, robot, intelligent, autonomous \\ \hline 
     \multirow{4}{*}{CTM} 
         & network, neural, learn, learning, artificial, evolutionary, language, recurrent, knowledge, processing \\ 
         & gene, expression, datum, model, analysis, microarray, cluster, clustering, genetic, classification \\ 
         & market, stock, price, return, risk, financial, rate, option, work, volatility \\
         & decision, agent, make, making, multi, human, group, uncertainty, social, approach \\ \hline
     \multirow{4}{*}{ProdLDA} 
         & network, neural, learn, recurrent, artificial, learning, evolutionary, language,  knowledge, adaptive  \\
         &  expression,  gene, datum, cluster, analysis, microarray, factor, bind, classification, site \\
         & market, stock, price, risk, financial, rate, evidence, return, exchange, work \\
         & decision, make, agent, making, group, environment, autonomous, robot, human, mobile \\ \hline
     \multirow{4}{*}{ETM} 
         & network, neural, gene, expression, datum, cluster, classification, recurrent, learn, genetic \\
         &  -  \\
         & market, gas, price, stock, financial, natural, return, work, rate, estimate \\
         & model, decision, base, analysis, method, theory, application, approach, make, dynamic   \\ \hline

     \multirow{4}{*}{LDA} 
        & network, neural, learn, learning, recurrent, dynamic, model, artificial, sensor, bayesian   \\
        &  gene, expression, datum, cluster, analysis, model, microarray, feature, sequence, base \\
        &  price, stock, oil, option, market, term, model, asset, return, pricing  \\
        & decision, theory, model, make, base, information, making, access, agent, bioinformatic
   \\ \hline

\end{tabular}
\end{adjustbox}
\caption{Some related topics discovered by different topic models in the M10 corpus when run for $T=10$ topics. \label{tab:topics_M10}}
\end{table*}

\begin{table*}
\centering
\begin{adjustbox}{width=1.0\linewidth}
  \begin{tabular}{ | c | c | } \hline
    \textbf{Model} & \textbf{Topics} \\ \hline

     \multirow{3}{*}{CTM-Neg} 
         & neural, network, recurrent, language, artificial, grammatical, context, learn, symbolic, natural \\ 
         & classification, neural, recognition, classifier, learn, pattern, coding, feature, network, sparse \\ 
         & network, neural, recurrent, feedforward, artificial, genetic, bayesian, learn, knowledge, evolutionary \\ \hline
     \multirow{1}{*}{LDA} 
        &  network, neural, recurrent, learn, \textit{mechanic}, adaptive, inference, compute, \textit{title}, \textit{extraction}  \\ \hline

\end{tabular}
\end{adjustbox}
\caption{Some AI-related topics discovered by CTM-Neg and LDA in the M10 corpus when run for $T=40$ topics. Italicized words in a topic appear less connected to the other words in the topic.   \label{tab:topics_M10_T40}}
\end{table*}

\subsection{Qualitative Evaluation}
It is acknowledged in the NLP community that automatic metrics do not always accurately capture the quality of topics produced by neural models \cite{hoyle2021automated}. So we perform manual evaluation of the topics for a few selected cases. 
Table \ref{tab:topics_20NG} shows some of the topics output by  random runs of the different topic models on 20NG for $T=20$ topics. Note that the table displays manually aligned topics, that is, the first topic mentioned against any of the topic models is very similar to the first topic stated against every other topic model, and similarly for all other topics.
We observe that the topics generated by CTM-Neg contain very specific words in the top positions that distinguish the topics more clearly compared to the case of other models. For example, the first topic produced by CTM-Neg contains very focused terms like `turkish', `israeli', `genocide', `war', etc., and is easily identifiable as `middle-east conflict' (corresponds to newsgroup \texttt{talk.politics.mideast} of 20NG corpus). 
CTM outputs a very similar topic but it seems to focus only on the `Armenian genocide' yet contains more generic terms like `neighbor' and `town'. ProdLDA also focuses primarily on `Armenian genocide' but its last word `jewish' probably refers to the Israeli conflict. While the corresponding topic from LDA  contains some generic terms like `man', `kill', etc., most of the words in ETM like `kill', `gun', and `fire' are very general. Moreover, words like `leave' and `start' that occur in this topic in ETM reduce the interpretability of the topic.

Similarly, the fourth topic in CTM-Neg is sports-related and contains specific words like `hockey' and `baseball'. While the corresponding topic from ProdLDA mentions  `hockey' (but not `baseball'), none of the other models produces these terms. The ability of CTM-Neg to extract focused words is probably a consequence of the negative sampling algorithm that encourages a topic to capture the most salient words of its representative documents so that deleting the topic pushes the reconstructed document away from the input document.

Table \ref{tab:topics_M10} shows the topics that are discovered in a random run of each topic model on the M10 dataset for $T=10$ topics. We show four topics -- the first is on `neural and evolutionary computing' (or `artificial intelligence'), the second on `microarray gene expression', the third on `stock market', and the fourth on `multi-agent decision making'. The topics generated by CTM and CTM-Neg are very similar. However, the  presence of words like `processing' in the first topic, `work' in the third topic, and `approach' in the fourth topic in CTM appear less connected to the other words in the respective topics. Such outliers are not visible in the topics produced by CTM-Neg. Moreover, the second topic output by CTM-Neg contains very domain-specific terms like `dna' and `motif', which are not produced by CTM. 
Similar issues occur in ProdLDA and LDA. In the case of ETM, the first topic contains words that make it a mixture of the first two topics produced by the other models. For example, it contains words like `neural' and `network' that occur in the first topic in the other models, and also `gene' and `expression' which are present in the second topic in the other models. Therefore, we have kept the second line for ETM topics in Table \ref{tab:topics_M10} blank. 
We observed that some of the topics produced by ETM contain many common words. 
In particular, we found that five topics from ETM contain the words `model', `decision', `method', `analysis', and `theory' in some order in the top slots, thus becoming repetitive, and consequently, ETM fails to discover meaningful and diverse topics like the other models. This is indicative of the component collapsing problem where all output topics are almost identical \cite{srivastava2017autoencoding}.

We have observed earlier that on the M10 corpus, for large topic counts LDA beats CTM-Neg in IRBO but not in NPMI. We revisit this issue now and manually analyze their topics for $T=40$. We found indeed the different topics output by LDA hardly overlap in words (leading to larger topic diversity) but the words do not always appear logically connected and interpretable (thus, sacrificing coherence). On the other hand, the topics generated by CTM-Neg look more coherent although they are not always disjoint. For example, see Table \ref{tab:topics_M10_T40} which shows the topics containing the word `neural' (among the top-10 words in the topic) discovered by CTM-Neg and LDA. CTM-Neg produces three topics that roughly relate to `natural language processing', `pattern recognition', and `neural and evolutionary computing', respectively. But only one topic from LDA contains `neural' -- it is primarily about `neural networks' but contains some very weakly related words.

\section{Conclusion}
We have proposed a negative sampling strategy for a neural contextualized topic model. We evaluated its performance on three publicly available datasets. In most of our experiments, the augmented model achieves higher topic coherence, as measured by NPMI and CV, and comparable topic diversity, as captured by IRBO, with respect to those of competitive topic models in the literature. A manual evaluation of a few selected topics shows that the topics generated by CTM-Neg are indeed coherent and diverse. In the future, we would like to compare it  with other contrastive learning-based topic models and integrate it with other neural topic models.

\section*{Acknowledgments}
This work is partially supported by the SERB-DST Project CRG/2021/000803 sponsored by the Department of Science and Technology, Government of India at Indian Association for the Cultivation of Science, Kolkata.

\appendix
\section{Appendix}
\label{sec:appendix}

\subsection{Detailed Results of Quantitative Evaluation}
Table \ref{Appendix:tab:Measures} shows the NPMI, CV, and IRBO scores obtained for the different topic models on the three datasets for different topic counts. This table has been used to construct Table \ref{tab:compareTM2} and Fig.  \ref{fig:compareTM} in this paper.

\begin{table*}[!ht]
    \centering
    \begin{adjustbox}{width=1\linewidth}
    \begin{tabular}{|cc|ccc|ccc|ccc|}
    \hline
     \multirow{2}{*}{\textbf{Model}}  & \multirow{2}{*}{\textbf{\#Topics}} & \multicolumn{3}{c|}{\textbf{Dataset: GN}} & \multicolumn{3}{c|}{\textbf{Dataset: 20NG}} & \multicolumn{3}{c|}{\textbf{Dataset: M10}} \\
     & & \textbf{NPMI} & \textbf{CV} & \textbf{IRBO} & \textbf{NPMI} & \textbf{CV} & \textbf{IRBO} & \textbf{NPMI} & \textbf{CV} & \textbf{IRBO} \\
    \hline
        CTM-Neg & 10 & -0.096 & 0.439 & 1 & 0.1 & 0.641 & 0.997 & 0.073 & 0.491 & 1 \\ 
        CTM-Neg & 20 & 0.031 & 0.424 & 1 & 0.129 & 0.672 & 0.994 & 0.076 & 0.49 & 0.993 \\ 
        CTM-Neg & 30 & 0.13 & 0.494 & 1 & 0.135 & 0.664 & 0.992 & 0.065 & 0.475 & 0.989 \\ 
        CTM-Neg & 40 & 0.174 & 0.542 & 0.999 & 0.129 & 0.656 & 0.991 & 0.06 & 0.461 & 0.986 \\
        CTM-Neg & 50 & 0.201 & 0.562 & 0.998 & 0.132 & 0.662 & 0.99 & 0.051 & 0.461 & 0.985 \\ 
        CTM-Neg & 60 & 0.227 & 0.592 & 0.998 & 0.125 & 0.65 & 0.991 & 0.046 & 0.449 & 0.982 \\ 
        CTM-Neg & 90 & 0.235 & 0.598 & 0.995 & 0.114 & 0.626 & 0.988 & 0.031 & 0.44 & 0.979 \\ 
        CTM-Neg & 120 & 0.234 & 0.592 & 0.993 & 0.107 & 0.616 & 0.986 & 0.019 & 0.426 & 0.976 \\ \hline
        CTM & 10 & -0.144 & 0.391 & 1 & 0.048 & 0.585 & 0.988 & 0.055 & 0.507 & 0.999 \\ 
        CTM & 20 & -0.01 & 0.406 & 0.998 & 0.105 & 0.649 & 0.994 & 0.06 & 0.467 & 0.988 \\ 
        CTM & 30 & 0.076 & 0.467 & 0.997 & 0.111 & 0.661 & 0.992 & 0.069 & 0.479 & 0.983 \\ 
        CTM & 40 & 0.126 & 0.515 & 0.996 & 0.104 & 0.644 & 0.991 & 0.051 & 0.466 & 0.98 \\ 
        CTM & 50 & 0.129 & 0.51 & 0.994 & 0.101 & 0.636 & 0.99 & 0.044 & 0.456 & 0.977 \\ 
        CTM & 60 & 0.149 & 0.519 & 0.993 & 0.095 & 0.628 & 0.99 & 0.042 & 0.452 & 0.974 \\
        CTM & 90 & 0.153 & 0.535 & 0.991 & 0.094 & 0.608 & 0.989 & 0.04 & 0.456 & 0.971 \\ 
        CTM & 120 & 0.17 & 0.538 & 0.99 & 0.086 & 0.606 & 0.986 & 0.025 & 0.443 & 0.968 \\ \hline
        ProdLDA & 10 & -0.103 & 0.431 & 1 & 0.069 & 0.602 & 0.986 & 0.047 & 0.477 & 0.999 \\ 
        ProdLDA & 20 & -0.007 & 0.403 & 1 & 0.09 & 0.645 & 0.991 & 0.017 & 0.45 & 0.992 \\ 
        ProdLDA & 30 & 0.029 & 0.463 & 0.999 & 0.088 & 0.637 & 0.99 & 0.026 & 0.456 & 0.987 \\ 
        ProdLDA & 40 & 0.081 & 0.491 & 0.997 & 0.087 & 0.63 & 0.993 & 0.035 & 0.458 & 0.982 \\ 
        ProdLDA & 50 & 0.071 & 0.47 & 0.996 & 0.083 & 0.607 & 0.993 & 0.013 & 0.426 & 0.981 \\ 
        ProdLDA & 60 & 0.098 & 0.481 & 0.995 & 0.085 & 0.607 & 0.991 & 0.028 & 0.447 & 0.979 \\
        ProdLDA & 90 & 0.136 & 0.508 & 0.992 & 0.068 & 0.573 & 0.991 & 0.017 & 0.437 & 0.975 \\ 
        ProdLDA & 120 & 0.14 & 0.518 & 0.99 & 0.068 & 0.57 & 0.99 & 0.02 & 0.436 & 0.969 \\ \hline
        ETM & 10 & -0.235 & 0.411 & 0.549 & 0.05 & 0.531 & 0.883 & -0.004 & 0.325 & 0.653 \\
        ETM & 20 & -0.233 & 0.416 & 0.473 & 0.054 & 0.534 & 0.812 & -0.07 & 0.346 & 0.525 \\ 
        ETM & 30 & -0.269 & 0.438 & 0.578 & 0.05 & 0.533 & 0.788 & -0.063 & 0.333 & 0.449 \\ 
        ETM & 40 & -0.289 & 0.422 & 0.652 & 0.048 & 0.527 & 0.782 & -0.061 & 0.353 & 0.462 \\ 
        ETM & 50 & -0.245 & 0.393 & 0.669 & 0.048 & 0.526 & 0.803 & -0.069 & 0.35 & 0.523 \\ 
        ETM & 60 & -0.285 & 0.417 & 0.676 & 0.044 & 0.518 & 0.797 & -0.058 & 0.351 & 0.437 \\ 
        ETM & 90 & -0.274 & 0.402 & 0.699 & 0.048 & 0.527 & 0.834 & -0.056 & 0.351 & 0.464 \\ 
        ETM & 120 & -0.277 & 0.415 & 0.722 & 0.047 & 0.525 & 0.851 & -0.069 & 0.353 & 0.505 \\ \hline
        LDA & 10 & -0.18 & 0.391 & 0.996 & 0.065 & 0.554 & 0.942 & -0.035 & 0.389 & 0.966 \\ 
        LDA & 20 & -0.167 & 0.368 & 0.998 & 0.099 & 0.613 & 0.977 & -0.133 & 0.373 & 0.986 \\ 
        LDA & 30 & -0.173 & 0.393 & 0.999 & 0.097 & 0.609 & 0.985 & -0.183 & 0.374 & 0.991 \\ 
        LDA & 40 & -0.171 & 0.404 & 0.998 & 0.095 & 0.605 & 0.989 & -0.218 & 0.389 & 0.993 \\ 
        LDA & 50 & -0.198 & 0.406 & 0.999 & 0.085 & 0.584 & 0.99 & -0.251 & 0.402 & 0.995 \\ 
        LDA & 60 & -0.21 & 0.411 & 0.999 & 0.075 & 0.571 & 0.993 & -0.261 & 0.408 & 0.994 \\ 
        LDA & 90 & -0.206 & 0.407 & 0.999 & 0.054 & 0.54 & 0.995 & -0.274 & 0.398 & 0.994 \\ 
        LDA & 120 & -0.013 & 0.44 & 0.989 & 0.028 & 0.494 & 0.996 & -0.184 & 0.352 & 0.99 \\ \hline
    \end{tabular}
    \end{adjustbox}
    \caption{Performance of the different topic models on GN, 20NG, and M10 datasets for different topic counts. Each score is the median of 5 independent runs.  \label{Appendix:tab:Measures}}
\end{table*}

\end{document}